\documentclass[letterpaper, 10 pt, conference]{ieeeconf}  % Comment this line out if you need a4paper
\renewcommand{\thefootnote}{\alph{footnote}}

\IEEEoverridecommandlockouts                              % This command is only needed if 
\usepackage{comment}
\usepackage{caption}
\usepackage{subcaption}
\usepackage{amsmath,amsfonts,amssymb}

\usepackage{array}
\usepackage{gensymb}
\usepackage{float}
\usepackage{textcomp}
\usepackage{graphicx}

\usepackage{amsmath}
\usepackage{amssymb}
\usepackage{mathrsfs}
\usepackage{cite}
\usepackage[colorlinks=true, allcolors=blue]{hyperref}
\usepackage{hyperref}

\title{\LARGE \bf
A  Null Space Compliance Approach for Maintaining Safety and \\Tracking Performance in Human-Robot Interactions
}

\author{Zi-Qi Yang$^{1}$,  Miaomiao Wang$^{2}$, and Mehrdad R. Kermani$^{1}$% <-this % stops a space
\thanks{$^{1}$Zi-Qi Yang and Mehrdad R. Kermani are with the Department of Electrical and Computer
Engineering, Western University, London, ON N6A 5B9, Canada.
        {\tt\small Email: zyang524@uwo.ca, mkermani@eng.uwo.ca}}%
\thanks{$^{2}$Miaomiao Wang is with the School of Artificial Intelligence and Automation, Huazhong University of Science and Technology, Wuhan 430074,
China. 
        {\tt\small Email: mmwang@hust.edu.cn}}%
}

\begin{document}

\maketitle
\thispagestyle{plain}
\pagestyle{plain}

\begin{abstract}
In recent years, the focus on developing robot manipulators has shifted towards prioritizing safety in Human-Robot Interaction (HRI). Impedance control is a typical approach for interaction control in collaboration tasks. However, such a control approach has two main limitations: 1) the end-effector (EE)'s limited compliance to adapt to unknown physical interactions, and 2) inability of the robot body to compliantly adapt to unknown physical interactions. In this work, we present an approach to address these drawbacks. 
We introduce a modified  Cartesian impedance control method combined with a Dynamical System (DS)-based motion generator, aimed at enhancing the interaction capability of the EE without compromising main task tracking performance. This approach enables human coworkers to interact with the EE on-the-fly, e.g. tool changeover, after which the robot compliantly resumes its task.
% transitions back to pre-defined tasks without programmatic interruptions of motion. 
Additionally, combining with a new null space impedance control method enables the robot body to exhibit compliant behaviour in response to interactions, avoiding serious injuries from accidental contact while mitigating the impact on main task tracking performance. Finally, we prove the passivity of the system and validate the proposed approach through comprehensive comparative experiments on a 7 Degree-of-Freedom (DOF) KUKA LWR IV+ robot. \footnote{Video of the experiments can be found in the supplementary material.}

\end{abstract}

\section{Introduction}
Industrial robots are increasingly attracting attention in sectors that require heavy labour due to their reliability, cost-effectiveness, and efficiency.  However, not all manufacturing processes, such as automotive assembly and electronics manufacturing, can be fully automated without human intervention. Recently, advancements in safe HRI have been achieved with robot manipulators through various approaches: exteroceptive sensor-based methods, i.e. proximity sensor \cite{navarro2021proximity}; 
% robotic skin \cite{cirillo2015conformable}; 
% structure/actuator-based approaches 
% \cite{haddadin2016physical, yang2023computationally}; 
%van2009compliant
actuator-based approaches 
\cite{yang2023computationally}; 
and software/controller-based solutions \cite{schumacher2019introductory}. 
%calanca2015review
Among these methods, software/controller-based solutions are widely adopted in safe HRI applications due to their low cost, high versatility, and ease of implementation.  Active compliant control, a method within this category, significantly contributes to safe HRI practices.
% Active compliant control, a method within this category, exemplifies this potential and is rapidly evolving, making substantial contributions to the ongoing advancements in safe HRI practices.

Active compliant control often involves techniques aimed at shaping the impedance or admittance of robots.
% \cite{calanca2015review}.  
These techniques are particularly valuable for ensuring safe HRI, especially when the robot is in contact with the environment \cite{hogan1985impedance}. 
%,haninger2022towards
Building upon this foundation, and recognizing the predominant use of Cartesian space control in industrial applications, Cartesian impedance control is introduced \cite{ott2008cartesian}.
\begin{figure} [H]
   \begin{center}
\includegraphics[height=4.7cm]{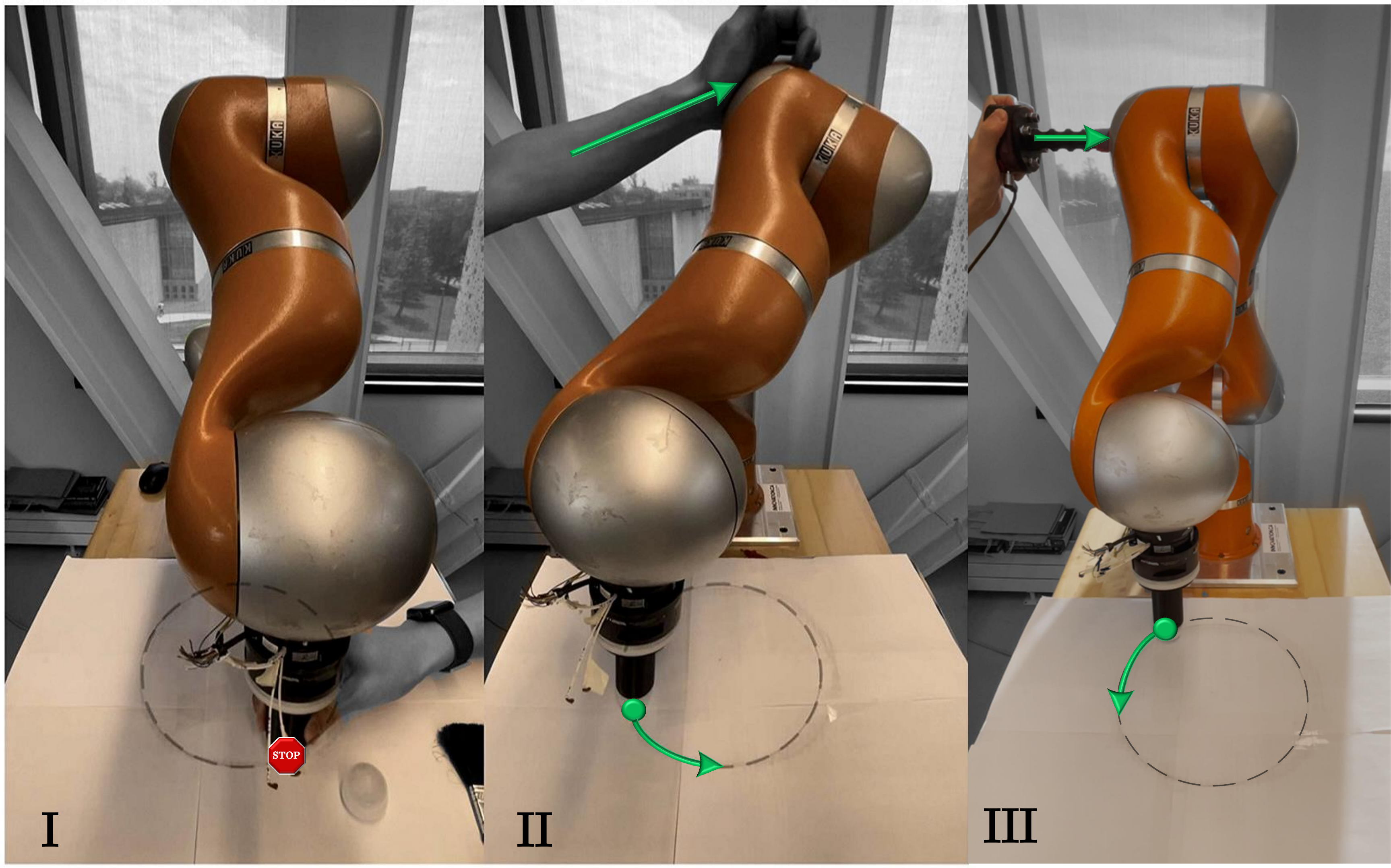}
   \end{center}
   \caption{Interaction with robot while (I) holding the EE, (II) interacting with the robot body without affecting tracking, and {(III) interacting using a handheld F/T sensor.}}
   \label{FirstPage}
\end{figure} 

Specifically, classical Cartesian impedance-controlled robots maintain compliant behaviour by allowing deviations from the desired path in response to environmental obstacles or external forces, with interaction forces regulated by impedance parameters. However, post-interaction motion can become jerky if these parameters are not appropriately chosen, due to significant correction forces caused by large positional discrepancies during interactions.
To address this issue, the correction force can be regulated by adjusting the impedance parameters through adaptive approaches \cite{abu2020variable}. However, these methods do not proactively change the robot’s trajectory to avoid unsafe interactions. In contrast, trajectory adaptation methods offer a more proactive solution by integrating trajectory adaptation with impedance control to ensure smooth transitions 
% \cite{luo2020adaptive, 
\cite{lin2021unified}. However, these approaches often introduce complexity and can cause undesirable delays during interactions. A more streamlined solution is to use a Dynamical System (DS)-based motion generator, which can instantly adapt to interactions without modifying the impedance characteristics \cite{khansari2014learning}. In \cite{khoramshahi2018human},
% \cite{amanhoud2019dynamical}, 
a state-dependent DS motion generator is introduced, enabling trajectory adaptation to human interactions.

Achieving compliant motion at the EE is essential, while unintentional contact issues can also arise on the robot body \cite{haddadin2017robot}. This issue is particularly significant when using kinematically redundant manipulators for human-robot collaboration tasks.
% \textcolor{red}{ such as learn from demonstration \cite{wang2018facilitating}.} 
To address this challenge, exploiting redundancy to enable compliant behaviour in the joint space is a potential solution. A common approach for redundancy control is the null space projection method. In the literature,  the null space motion can be regulated through optimization to guarantee the natural behaviour of robots during co-manipulation \cite{ficuciello2015variable}, or by hierarchical control to allow null space motion to resolve conflicts with higher priority tasks \cite{dietrich2019hierarchical}.  However, addressing the impact of unknown external interaction forces exerted on the robot body remains challenging. 
In \cite{sadeghian2013task}, a null space impedance control is employed to react to the interaction forces acting on the robot body, complemented by an observer to compensate for Cartesian space errors. Nonetheless, this approach relies on accurate knowledge or estimation of the robot's inertia matrix, which may not always be feasible in real-world settings. 

To address these challenges, we aim to develop a new method that enables the robot to perform pre-defined tasks while exhibiting compliant behaviour to unknown interactions along the robot structure, ensuring tracking performance is unaffected without requiring the measurement or estimation of interaction forces and robot dynamics information.
% the robot manipulator to execute a pre-defined task while rendering compliant behaviour along the robot structure to unknown interactions, and 2) ensures that the tracking performance of the task remains unaffected by unknown interactions exerted on the robot structure, without requiring the measurement or estimation of external torques and robot dynamics information.}}} 
Our method enables the main task tracking in non-interaction situations while also allowing unknown interventions to interrupt EE motion, such as for tool changeover, then smoothly return to the desired path, as shown in Fig. \ref{FirstPage} (I). To achieve these, we introduce an enhanced Cartesian impedance control method combined with a DS-based motion generator to generate interactable EE motions on-the-fly.
% without programmatic interruptions of motion.
Furthermore,  we propose a novel null space impedance control strategy with joint friction torque compensation. This method efficiently extends compliant behaviour to the null space of the pre-defined main task, ensuring the tracking performance of the main task remains unaffected under unknown external interactions exerted on the robot body, as shown in Fig. \ref{FirstPage} (II).
The main contributions of this work are summarized as:
\begin{enumerate}
  \item 
  Introducing a modified Cartesian impedance control method combined with a DS-based motion generator, enhancing robot compliance during unknown physical interactions at the EE. This method ensures main task tracking and enables post-interaction path re-planning, allowing the EE to transition smoothly between interactions and the main task.

  \item Introducing a novel null space impedance control method that enables the robot body to compliantly respond to unknown external interactions, without compromising the tracking performance of the main task. This method effectively dissipates external interaction energy in the null space, preserving tracking performance without requiring measurement or estimation of external forces, and robot dynamics information.
  
  \item Demonstrating the proposed method's passivity under conditions of unknown interaction forces and robot dynamics. This theoretically validates the effectiveness and safety of the proposed method for use in complex real-world HRI applications.

  % \item \textcolor{red}{TBD}Conducting a series of comparative experiments on a 7-DOF KUKA LWR IV+ robot, and assessing the effectiveness of our proposed control scheme compared to the existing methods in the literature.
\end{enumerate}
The proposed control scheme is evaluated under conditions of unknown interaction forces and robot dynamics. The results highlight its ability to achieve accurate, repeatable, interactable, interruptible motion and robot compliance. 

The rest of this paper is organized as follows: Section \ref{control objective} outlines the control objectives. Section \ref{Methods} details the proposed control method and proves the passivity of the system. Section \ref{Experimental Evaluation} {presents a series of comparative experiments and provides a numerical comparison.} {Section \ref{Discussion} discusses additional
experimental findings, future directions, and concludes this work.}

% \section{Problem statement}
% \label{Problem statement}
% In industrial applications, pre-programmed Cartesian paths with specific impedance parameters are utilized to achieve low-level compliance within the Cartesian space, allowing for temporary adaptation to obstacles and interactions. To extend this compliance capability to HRI tasks, we aim to develop a method that {\textit{ 1) enables the robot EE to accurately track a desired Cartesian space path and react compliantly to interactions with coworkers, and 2) effectively dissipates energy from unknown external torques exerted on the robot body, thereby ensuring that Cartesian space path tracking remains unaffected without the need for additional F/T sensors.}}

\section{Control Objective}
\label{control objective}
In this work, we consider an $n$-DOF redundant serial robot manipulator. The dynamic model is expressed as,
\begin{equation}
\label{dynamicmodel}
M(q)\ddot q + C(q,\dot q)\dot q + G(q) = \tau_{total} +\tau^f + \tau^{ext},
\end{equation}
where ${q \in \mathbb{R}^{n } }$ denotes the joint configuration, ${M(q) \in \mathbb{R}^{n \times n} }$ is the inertia matrix.  ${C(q,\dot q) \in \mathbb{R}^{n \times n} }$ represents the Coriolis and centrifugal matrix, and ${G(q) \in \mathbb{R}^{n} }$ denotes the gravitational torques. ${\tau_{total} \in \mathbb{R}^{n} }$ denotes the total joint torque to be designed,  ${\tau^{f} \in \mathbb{R}^{n} }$ denotes the joint frictional torque, and ${\tau^{ext}\in \mathbb{R}^{n }}$ denotes the unknown external interaction torque exerted on the robot joint. In the following, the dependency on joint configuration $(q, \dot q)$ is omitted for the sake of brevity. %Moreover, it is assumed that Coriolis and centrifugal torques are negligible due to the slowly varying joint velocities.

The total control torque $\tau_{total}$ in \eqref{dynamicmodel} can be designed as $\tau_{total} = G(q)+\tau^c + \tau^n$,
% \begin{equation}
% \label{tau total}
% \tau_{total} = G(q)+\tau^c + \tau^n,
% \end{equation}
where ${\tau^{c} \in \mathbb{R}^{n}}$ represents the Cartesian space joint control torque,  ${\tau^{n} \in \mathbb{R}^{n} }$ represents the null space joint control torque, and $G(q)$ is the gravitational torque defined previously. It should be noted that the effect of gravitational torque can be accurately compensated in most industrial robot's internal controllers.

% \begin{figure*} [t]
%    \begin{center}
% \includegraphics[width=11cm,height=4cm]{Fig/Control_Diagram.pdf}
%    \end{center}
% \caption{Control Diagram. (I) Cartesian motion generation. (II) Tracking control. (III) Modified Cartesian impedance control. (IV) Proposed joint friction compensation. (V) Proposed null space impedance control.}
%    \label{Control_Diagram}
% \end{figure*} 

{Let
$x_d=[p_d \; r_d]^T$, \text{and} $x=[p \; r]^T\in \mathbb{R}^6$ denote the desired and measured poses of the EE in 6D Cartesian space, in that $p_d = [p^x_d, p^y_d, p^z_d]^T$, \text{and} $p = [p^x, p^y,p^z]^T\in\mathbb{R}^3$ are the desired and measured positions, and $ r_d = [\phi_d, \theta_d,\psi_d]^T$, \text{and} $ r = [\phi, \theta,\psi]^T  \in \mathbb R^3$ are the desired and measured orientations of the EE, respectively. Considering the total number of samples, $t$, each sample point of each vector is represented as \(p_{d,i}^x\) and \(p_i^x\) where \( i \in \{1,2,\dots,t\} \). }

{In this work, our primary objectives are to: 
% 1) maintain the tracking performance of the main task (e.g. $\| x_d-x \|$) under both non-interaction and interaction scenarios, } 
% 1) maintain tracking performance (e.g., $\| x_d-x \|$) under the interaction scenarios to be equal or less than that in the non-interaction scenario, }
1) maintain the tracking performance during interactions with the robot body, defined as,
\begin{align}
\label{NMAE}
\frac{\max\limits_{i=1}^{t}|p_{d,i}^x-p^{x}_i|}{\max\limits_{i=1}^{t}p_{d,i}^x - \min\limits_{i=1}^{t}p_{d,i}^x},
\end{align}
along the $x-$axis, as well as those for the $y-$ and $z-$ axes to remain at or below the level observed during the non-interaction scenario,}
2) enable compliant behaviour of both the EE and the robot body when reacting to unknown external interactions, and 3) achieve objectives 1) and 2) without relying on the measurement or estimation of external interaction forces, as well as robot dynamics information. {Details of non-interactive and interactive scenarios are given in Section \ref{Experimental Evaluation}}.

% \begin{enumerate}
%   \item To reduce Cartesian space tracking error $ \| x_d-x \|$, 
%   ensuring both translational and rotational tracking accuracy of the EE in non-contact situations.
%   \item To reduce $ \| x_d - x \|$ under the effect of unknown external torques exerted on the robot body by the user or obstacle.  
%   \item \textcolor{red}{To limit the force applied to the user during EE interactions, such as during tool changeovers. }
% \end{enumerate}

\begin{figure} [!ht]
   \begin{center}
\includegraphics[width=8.5cm,height=3.25cm]%width=7cm,height=3cm
{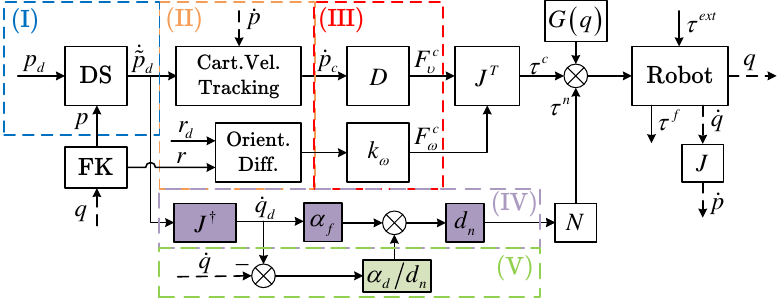}
   \end{center}
   \caption{Control Diagram: (I) Cartesian motion generation, (II) tracking control, (III) modified Cartesian impedance control, (IV) proposed joint friction compensation, and (V) proposed null space impedance control.}
   \label{Control_Diagram}
\end{figure} 
\section{Methods}
\label{Methods}
In this section, we present our control architecture in Fig. \ref{Control_Diagram}. The components include a Cartesian motion generator, tracking controllers, a modified Cartesian impedance controller,  a proposed
joint friction compensation approach, and a proposed null space impedance control strategy.

\subsection{Cartesian Space Motion Generation}
\subsubsection{Motion Generator}
\label{Cartesian Space Motion Generation}
In this subsection, we introduce a continuously differentiable state-dependent DS-based function $g(\mathscr{Z})$ in a feedback configuration, inspired by \cite{khoramshahi2018human, kronander2015passive}.
% khansari2011learning
Here, \( \mathscr{Z}:=[\zeta_1,...,\zeta_s]^T \) represents a set of state variables containing desired and measured Cartesian positions or orientations of the EE with \( \zeta_i \in \mathbb{R}^{3} \) and \( i \in \{1,2,\dots,s\} \). %\textcolor{red}{These variables can contain desired and measured EE Cartesian positions or orientations.} 
First, the DS stores the translational component  $p_d$ of the desired Cartesian pose $x_d$
% ${ p_d = [p^x_d, p^y_d, p^z_d]^T \in \mathbb{R}^3 }$ 
% of the desired Cartesian pose $x_d=[p_d \mid r_d]^T \in \mathbb R^{6}$
of a pre-defined task. The rotational component $r_d$ 
will be addressed in subsection \ref{Cartesian impedance control}.  
Second, to accommodate task-specific requirements, the  stored $p_d$ can be transformed to the desired position using  a function $g: \mathbb{R}^3 \times \mathbb{R}^3 \rightarrow \mathbb{R}^3$ describing the DS,%, to generate a transformed desired position 
\begin{equation}
\label{tilde pd 1}
{\tilde{p}}_d = g(p_d,p; \Lambda),
\end{equation}
where % $g: \mathbb{R}^3 \times \mathbb{R}^3 \rightarrow \mathbb{R}^3$ represents a function describing the DS,
$p$ is obtained from the forward kinematics (FK), and \( \Lambda= [ \lambda_1, \dots, \lambda_t ]^T \), with \( \lambda_i \in \mathbb{R}^+ \) and \( i \in \{1,2,\dots,t\} \), denotes the set of independent parameters utilized for state variable transformation. 
In the context of repetitive tasks, $\Lambda$ can be regarded as fixed parameters. Thirdly, taking the derivative of the desired position \eqref{tilde pd 1},  the desired translational velocity $\dot{\tilde{p}}_d=[\dot{\tilde{p}}_d^x, \dot{\tilde{p}}_d^y, \dot{\tilde{p}}_d^z]^T \in \mathbb R^3$ can be expressed as
\begin{equation}
\label{dot tilde pd}
\dot{\tilde{p}}_d  = \frac{d}{dt} g({p}_d,p;\Lambda). %\dot g({p}_d,p;\Lambda),
\end{equation}
%where $\dot{\tilde{p}}_d=[\dot{\tilde{p}}_d^x, \dot{\tilde{p}}_d^y, \dot{\tilde{p}}_d^z]^T \in \mathbb R^3$. 
% Here, $\tilde{p}_d=[{\tilde{p}}_d^x,
% {\tilde{p}}_d^y,
% {\tilde{p}}_d^z]^T$ represents the transformed desired EE position, 
A specific application of \eqref{tilde pd 1} and \eqref{dot tilde pd} for a circle drawing task is provided in the experiment section. Note that in our design $\dot{\tilde{p}}_d$ serves as the reference input signal for the Cartesian velocity tracking controller as shown in Fig. \ref{Control_Diagram} (II).  In this setup, we utilize  a Proportional--Derivative (PD) velocity tracking controller to ensure  tracking performance,
\begin{equation}
\label{dot pc}
\dot p_c = k_p (\dot{\tilde{p}}_d - \dot{p}) + k_d (\ddot{\tilde{p}}_d - \ddot{p}),
\end{equation}
where $\dot p_c = [\dot{p}^x_c, \dot{p}^y_c, \dot{p}^z_c]^T \in \mathbb{R}^3$ represents the controlled output of the Cartesian velocity tracking controller, $\dot{p}=[\dot{p}^x, \dot{p}^y, \dot{p}^z]^T \in \mathbb{R}^3$ represents the measured EE translational velocity. The term $\ddot{\tilde{p}}_d$ can be obtained by applying numerical differentiation to $\dot{\tilde{p}}_d$. Here, $k_p \in \mathbb R^+$ and $k_d \in \mathbb R^+$ represent the user-defined proportional and the derivative gain, respectively.

\subsubsection{Cartesian impedance control }
\label{Cartesian impedance control}
In the previous subsection, we enabled the tracking of a task path by generating the desired translational velocity and implementing closed-loop velocity tracking. Based on this, we introduce a modified impedance controller to enable compliant behaviour. This controller derives the control torque from the result $\dot p_c$ of the Cartesian velocity tracking controller in \eqref{dot pc}.

Let $F^c=[F_\upsilon ^c \; F_\omega ^c]^T \in \mathbb R^{6}$ denote the Cartesian control force to be designed. The translational component of $F^c$, denoted by $F_\upsilon^c \in \mathbb R^3$, can be designed as,
\begin{equation}\label{Fcv}
F_\upsilon^c  = D\:\dot p_c
\end{equation}
where {$D=\operatorname{diag}(d_\upsilon^{x}, d_\upsilon^{y}, d_\upsilon^{z})$, with $d_\upsilon^{x}, d_\upsilon^{y}, d_\upsilon^{z} \in \mathbb R^+$}, denotes a diagonal damping matrix for translational motion in Cartesian space with positive entries. By reducing $D$, the EE can respond more compliantly to contacts from users or obstacles. This smooth response enables human coworkers to perform interactions (e.g., tool changeover) without requiring programmatic interruptions of motion. 

The preceding focused on the translational motion to simplify the discussion. However, tasks such as polishing, which require maintaining the tool perpendicular to the surface, necessitate addressing the orientation control of the EE.  The orientation control force $F_\omega^c   \in \mathbb R^3$ can be formulated as,
\begin{equation}
\label{Fcw}
   F_\omega^c = K_\omega( r_d -  r),
\end{equation}
% where $ r_d = [\phi_d, \theta_d,\psi_d]^T  \in \mathbb R^3$ represents the desired EE orientation, $ r = [\phi, \theta,\psi]^T  \in \mathbb R^3$ represents the measured EE orientation, 
where {$K_\omega=\operatorname{diag}(k_\omega^{\phi}, k_\omega^{\theta}, k_\omega^{\psi})$, with  $k_\omega^{\phi}, k_\omega^{\theta}, k_\omega^{\psi} \in \mathbb R^+$,} represents a diagonal stiffness matrix for the orientational motion of the EE with positive entries. Subsequently, the control torque $\tau^c \in \mathbb R^n$ can be expressed as,
\begin{equation}
\label{tauc}
\tau^c = J^T F^c,
\end{equation}
where $J \in \mathbb R^{6\times n}$ represents the Jacobian
matrix of the $n$-DOF robot manipulator with a full rank of 6.  
%Note that in (\ref{Fcw}), it is necessary to convert the Euler angles to Quaternion representation to avoid discontinuities when calculating the orientation error.  Further discussion regarding the orientation tracking accuracy is included in the experimental evaluation section.
Note that a more popular way to compute the orientation error in \eqref{Fcw} is to convert the Euler angles to Quaternion representation. Further discussion regarding the orientation tracking accuracy is provided in \hyperref[A]{Exp. A}, as shown in Fig. \ref{no_contact_tracking+boxplot} (c).

\subsection{{Null Space Impedance Control}}
\label{Null Space Impedance Control}
% \subsubsection{Controller Design}
With the methods discussed in the previous subsections, we have enabled the EE to track the task path and safely interact with the users. This subsection focuses on allowing the robot body to respond to external interactions compliantly by controlling redundant DOFs, to reduce interference with Cartesian space motion. In other words,  external energy exerted on the robot body can be effectively dissipated in the null space.

Let the relation between measured EE velocity $\dot x \in \mathbb R^6$ and measured joint velocity $\dot q \in \mathbb R^n$ be expressed as $\dot x = J \dot q$.
% \begin{equation}
% \label{dot x}
% \dot x = J \dot q.
% \end{equation}
A redundancy solution ($n > 6$) to it is achieved through a null space projector $\dot q = J^\dagger \dot x + N \dot q$,
% \begin{equation}
% \label{dot q}
% \dot q = J^\dagger \dot x + N \dot q,
% \end{equation}
where $J(q)^\dagger = J(q)^T(J(q)J(q)^T)^{-1} \in \mathbb R^{n\times 6}$ denotes the pseudoinverse of $J(q)$. $N \in \mathbb R^{n\times n}$ is a null space projection matrix of $J(q)$, denoted as $N=I-J^\dagger J$,
% \begin{equation}
% \label{N}
% N=I-J^\dagger J,
% \end{equation}
where $I \in\mathbb R^{n\times n}$ is an identity matrix.  By utilizing the null space projection, redundant DOFs can be exploited to move within the null space of $J(q)$.
This implies that the projection can be potentially used to project the external interaction forces to the null space. 
This is particularly valuable in scenarios where the robot interacts with its environment.
% , such as in the HRC settings \cite{zhang2024hierarchical}.
In light of this, we propose a novel null space impedance control law formulated as,
\begin{equation}
\label{taun}
  \tau^n = N d_n (\alpha_f \dot q_{d} + \frac{\alpha_d}{d_n}(\dot q_{d} - \dot q)),
\end{equation}% -\tau^{ext}
where $d_n \in \mathbb R^+$ represents the null space damping parameter. $\dot q_{d} = J(q)^\dagger \dot{\tilde{x}}_d
\in \mathbb R^{n}$ denotes the desired joint velocity converted from the transformed desired velocity of the EE, denoted as $\dot{\tilde{x}}_d=[\dot{\tilde{p}}_d\mid\dot r_d]^T \in \mathbb R^{6}$. In practical implementation, the desired orientation velocity $\dot r_d$ of $\dot{\tilde{x}}_d$ is set to zero to maintain matrix integrity, with the orientation of the EE controlled by (\ref{Fcw}). The first term, $Nd_n \alpha_f \dot q_{d}$, functions as a feedforward compensation term for joint frictional torques, where $\alpha_f$ denotes the frictional gain parameter. By using the desired joint velocity \(\dot{q}_d\), which provides a cleaner signal, and the null space projector \(N\), the compensation is applied smoothly to the redundant DOFs. This enhances the joint responsiveness to external interactions.
The second term $N \alpha_d(\dot q_{d} - \dot q)$ utilizes null space projection to extend the joint motion into the null space of Cartesian space main tasks (e.g. polishing, buffing). Here, \( \alpha_d \) represents the damping gain parameter that regulates the level of compliance. This term enables the redundant DOFs to respond compliantly to unknown external torque $\tau^{ext}$. The deviation between \( \dot{q} \) and \( \dot{q}_d \) caused by the external force generates damping torques through this term. The damping effect effectively dissipates external energy, reducing the impact on main task tracking performance.  By combining these two terms, we ensure that external energy is exclusively dissipated through the joint damping effect, without being expended in resisting joint friction. 

% Overall, the methods ensure the tracking performance of the main task while enabling compliant behaviour of the EE and the robot body when reacting to unknown interactions.

\subsection{Passivity Analysis}
In this subsection, we analyze the passivity of the overall control system. We introduce a storage function consisting of three sub-system storage functions as follows: $S = S_1 + S_2 +S_3$,
% \begin{align}
% \label{S}
% S &= S_1 + S_2 +S_3, 
% %&=\frac{1}{2} \dot{e}_q^T M \dot{e}_q + \frac{1}{2} {e}_r^T k_\omega  {e}_r
% %+ \frac{1}{2} k_d \dot{e}_p^T D \dot{e}_p,
% \end{align}   
where  $S_1=\frac{1}{2} \dot{e}_q^T M \dot{e}_q$ represents the kinetic energy associated with the joint velocity error induced by external interaction forces exerted on the robot body, $S_2=\frac{1}{2} {e}_r^T K_\omega  {e}_r$ represents the spring potential energy of the orientation controller, and $S_3=\frac{1}{2} k_d \dot{e}_p^T D \dot{e}_p$ represents the energy dissipated due to the damping effect in the Cartesian impedance controller. Moreover, $\dot e_q = \dot {q}_d - \dot q$ denotes the joint velocity error, $e_r = {r_d} - r$ represents the EE orientation error, $\dot e_p = \dot{\tilde{p}}_d -\dot{p}$ represents the EE translational velocity error.

Take the time derivative of $S$, we have,
\begin{align}
\label{dot S}
    \dot S = \dot S_1 + \dot{e}_r^T K_\omega {e}_r + k_d\dot{e}_p^T D \ddot{e}_p,
\end{align}
where $\dot S_1$ is given by,
\begin{equation}
\label{dot S1,1}
   \dot S_1 = \dot{e}_q^T M \ddot{e}_q + \frac{1}{2} \dot{e}_q^T \dot M \dot{e}_q.
\end{equation}

In HRI applications, due to safety requirements, task paths typically exhibit slow variations ($\ddot{q}_d, \dot{q}_d \approx 0$).  As a result, the contribution of Coriolis and centrifugal effects are negligible w.r.t. the external force, while inertial forces are insignificant w.r.t. that of the gravitational forces \cite{ficuciello2015variable,caccavale2003tricept}. Therefore, we assume  \(M\ddot{q}_d + C\dot{q}_d \approx 0\). Consequently, $M\ddot{e}_q$ in (\ref{dot S1,1}) can be decomposed and combined with (\ref{dynamicmodel}) to be written as $     M\ddot{e}_q = M\ddot{q}_d -M \ddot{q}= -C\dot{e}_q - \tau^c - \tau^n -\tau^f - \tau^{ext}$.
% \begin{align}
%      M\ddot{e}_q & = M\ddot{q}_d -M \ddot{q} \nonumber \\
%      %& = -C\dot{q}_d + C\dot q + G - \tau_{total} -\tau^f - \tau^{ext} \nonumber\\
%      % & = -C\dot{e}_q + G - (G+\tau^c + \tau^n) -\tau^f - \tau^{ext} \nonumber\\
%      & = -C\dot{e}_q - \tau^c - \tau^n -\tau^f - \tau^{ext}.
% \end{align}

Substituting the above expression into (\ref{dot S1,1}), and consider skew symmetry of $(\dot M-2C)$, we have,
\begin{align}
   \dot S_1 
   % &= \dot{e}_q^T (-C\dot{e}_q - \tau^c - \tau^n -\tau^f - \tau^{ext})+ \frac{1}{2} \dot{e}_q^T \dot M \dot{e}_q\\
   &= \frac{1}{2}\dot{e}_q^T (\dot M-2C)\dot{e}_q- \dot{e}_q^T(\tau^c + \tau^n +\tau^f + \tau^{ext})\nonumber \\
   &= - \dot{e}_q^T(\tau^c + \tau^n +\tau^f + \tau^{ext}).  
   \label{dot S1,2}
\end{align} 

By substituting (\ref{dot pc})% (\ref{Fcv}), (\ref{Fcw}) 
--(\ref{tauc}) into (\ref{dot S1,2}), and considering  $\dot{x}=[\dot{p} \mid \dot r]^T \in \mathbb R^{6}$, we obtain,
\begin{align}
\dot S_1 
    &= - \dot{e}_q^T  J^T \begin{bmatrix}
    F^c_v\\
    F^c_w
   \end{bmatrix} - \dot{e}_q^T(\tau^n +\tau^f + \tau^{ext}) \nonumber \\ 
   % &= - \left(J(\dot{e}_q)\right)^T \begin{bmatrix}
   %  F^c_v\\
   %  F^c_w
   % \end{bmatrix} - \dot{e}_q^T(\tau^n +\tau^f + \tau^{ext}) \\ 
   % &= - (\dot{\tilde{x}}_d -\dot{x})^T\begin{bmatrix}
   %  F^c_v\\
   %  F^c_w
   % \end{bmatrix} - \dot{e}_q^T(\tau^n +\tau^f + \tau^{ext})\\
   % &= -\begin{bmatrix}
   %     (\dot{\tilde{p}}_d -\dot{p})^T  &
   %     (\dot{r}_d -\dot{r})^T
   % \end{bmatrix}\begin{bmatrix}
   %  F^c_v\\
   %  F^c_w
   % \end{bmatrix} - \dot{e}_q^T(\tau^n +\tau^f + \tau^{ext})\\
   % &= -((\dot{\tilde{p}}_d -\dot{p})^T F_v^c  +
   %     (\dot{r}_d -\dot{r})^T F_\omega^c)- \dot{e}_q^T(\tau^n +\tau^f + \tau^{ext})\\
    % &= -\dot{e}_p^T D \dot p_c -
    %    \dot{e}_r^T k_\omega {e}_r - \dot{e}_q^T(\tau^n +\tau^f + \tau^{ext})\\
    % &=-\dot{e}_p^T D ( k_p \dot{e}_p + k_d \ddot{e}_p) - \dot{e}_r^T k_\omega {e}_r - \dot{e}_q^T(\tau^n +\tau^f + \tau^{ext})\\    
    &= -k_p \dot{e}_p^T D \dot{e}_p - k_d\dot{e}_p^T D \ddot{e}_p - \dot{e}_r^T K_\omega {e}_r \notag \nonumber \\
    &\quad - \dot{e}_q^T(\tau^n +\tau^f + \tau^{ext}) \nonumber \\
    %\label{dot S1 cancel}
    &\leq - k_d\dot{e}_p^T D \ddot{e}_p - \dot{e}_r^T K_\omega {e}_r - \dot{e}_q^T(\tau^n +\tau^f + \tau^{ext}), \label{dot S1}
\end{align} 
where  the quadratic term $ \dot{e}_p^T D \dot{e}_p$ is non-negative.
Combine (\ref{dot S}) with (\ref{dot S1}), we can cancel out the first two terms, i.e., 
\begin{align}
\label{dot S simplify}
    \dot S \leq - \dot{e}_q^T(\tau^n +\tau^f + \tau^{ext}). 
\end{align}

Substitute (\ref{taun}) into (\ref{dot S simplify}), and assume the friction torques can be approximately calculated by $-Nd_n \alpha_f \dot q$, we have,
\begin{align} 
   \dot S &\leq - \dot{e}_q^T (N d_n (\alpha_f \dot q_{d} + \frac{\alpha_d}{d_n}(\dot q_{d} - \dot q))+\tau^f) - \dot{e}_q^T \tau^{ext}
   \nonumber \\
   % &= - \dot{e}_q^T (N d_n \alpha_f \dot q_{d} - N d_n \alpha_f \dot q + N \alpha_d \dot e) - \dot{e}_q^T \tau^{ext}
   % \\  
   & =  - (d_n \alpha_f +  \alpha_d)  \dot{e}_q^T N  \dot{e}_q - \dot{e}_q^T \tau^{ext},  \label{dot S final}
\end{align}
where the quadratic term $\dot{e}_q^T N \dot{e}_q$ is non-negative. The terms $d_n \alpha_f \geq 0$ and $\alpha_d \geq 0$ represent the coefficients of dissipated energy due to joint friction and the damping effect introduced by the null space impedance control law, respectively. Together, these terms facilitate energy dissipation within the system. Finally, we can write,
$\dot S \leq - \dot{e}_q^T \tau^{ext}$,
% \begin{align}
% \label{S passivity}
%    \dot S \leq - \dot{e}_q^T \tau^{ext},
% \end{align}
which ensures the passivity of the system w.r.t. the input-output pair $(-\tau^{ext}, \dot{e}_q)$. To evaluate the proposed methods, we discuss experiments in the following section.

\section{Experimental Evaluation}
\label{Experimental Evaluation}
\begin{figure} [H]
   \begin{center}
\includegraphics[height=4.5cm]{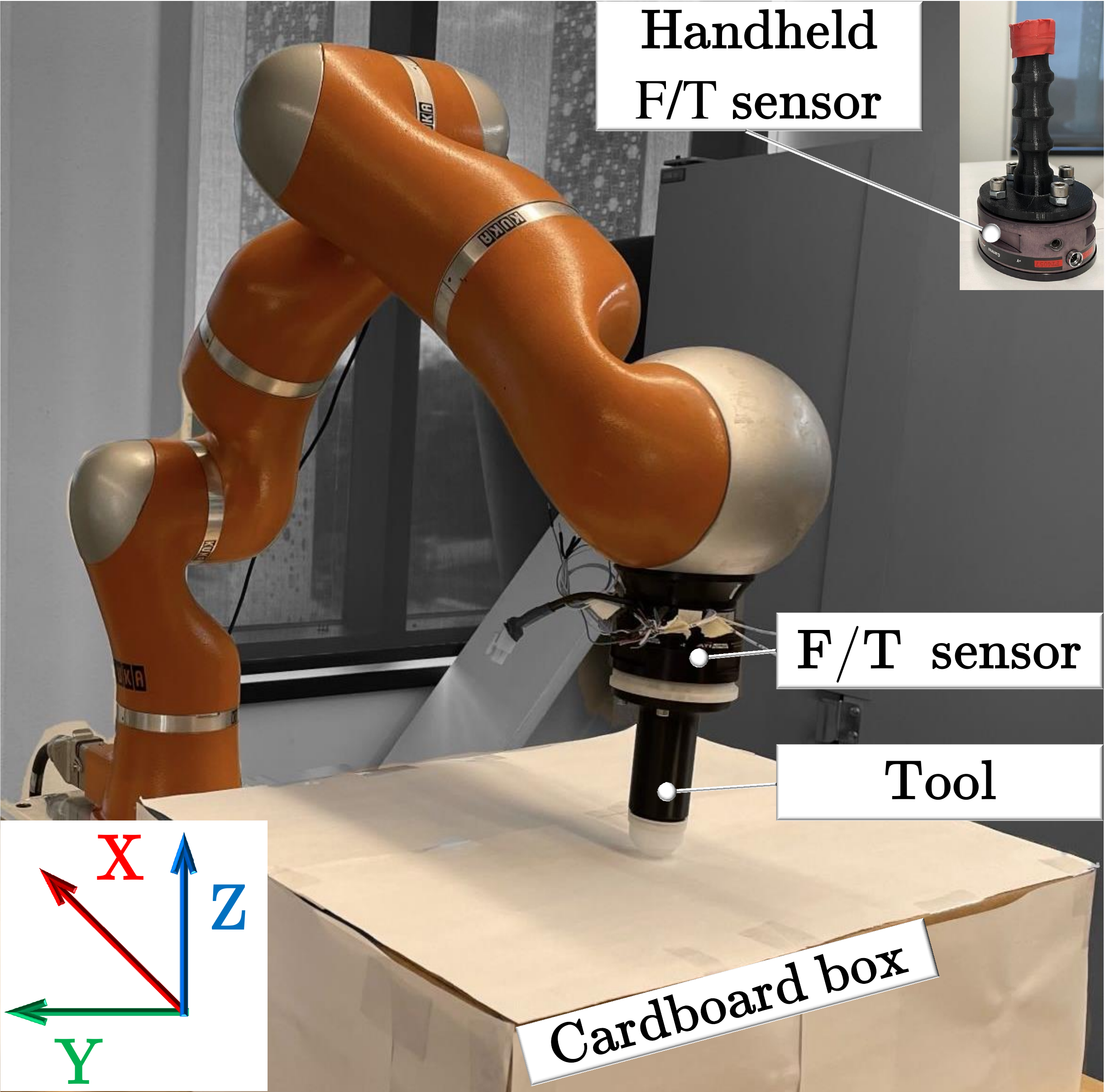}
   \end{center}
   \caption{{Experimental setup: 7-DOF robot arm with a 3D-printed tool,  a mounted and a handheld F/T sensor for measuring interactive forces.}}
\label{Experiment_Setup}
\end{figure}

The experiments are conducted on a 7-DOF KUKA LWR IV+ robot manipulator as shown in Fig. \ref{Experiment_Setup}. 
The control algorithms are implemented on a remote Ubuntu PC with the Robot Operating System (ROS) framework. This remote PC establishes communication with the KUKA robot controller via UDP, utilizing the Fast Research Interface (FRI) with a sampling rate of 200 Hz. A $6$--axis ATI Gamma F/T sensor is attached to the EE for reference purposes. Throughout the experiments, the robot is controlled in joint impedance mode, and proper compensation for gravitational torques is ensured, including the effects of the F/T sensor and a lightweight 3D-printed tool. {A handheld device incorporating the same ATI F/T sensor is used in robot body interaction experiments for reference purposes.}

To evaluate the proposed method, we consider a circle-drawing task to represent a typical polishing pattern. This task involves tracing a circle of radius \(\epsilon \in \mathbb{R}^+\) on the top flat surface of a cardboard box, centred at \(O=[a,b,0]^T \in \mathbb{R}^3\) in Cartesian coordinates. {The path is selected by the user to avoid kinematic singularities and joint limits. } To accommodate task variations, the transformed desired position is,
\begin{equation}
\label{tilde pd real}
{\tilde{p}}_d = {g}(p_d, p; \Lambda) := {\Bar{\lambda}_1}
\begin{bmatrix}\rho  \cos(\delta) -a  \\ \rho  \sin(\delta) - b  \\0\end{bmatrix}+\begin{bmatrix}
a_d \\ b_d\\ {\Bar{p}_d^z}
\end{bmatrix}.
\end{equation}
In this context, $\Bar{\lambda}_1 = \operatorname{diag}[\lambda_{1},\lambda_{1},1]$, where $\Lambda = \lambda_1 =\frac{\epsilon_{d}}{\epsilon} \in \mathbb{R}^+$, represents the scaling factor. Here, $\epsilon_d $ denotes the radius of the desired circle.  Consequently, the transformed position is represented in polar coordinate, where $\rho=\sqrt{({\Delta p^x})^2+({\Delta p^y})^2}$ and $\delta = atan2(\Delta p^y,\Delta p^x) \in [-\pi,\pi]$.  Here, $\Delta p =  [\Delta p^x, \Delta p^y, \Delta p^z]^T  \in \mathbb R^3$ denotes the transformed position errors along the $x$--axis, $y$--axis, and $z$--axis, respectively. Specifically, $\Delta p$ is defined as $\Delta p = \Bar{\lambda}_1^{-1}(p-O_d)$, where $O_d = [a_d,b_d,\Bar{p}_d^z]\in \mathbb{R}^3$ represents the translated center of the circle. 
Subsequently, the transformed desired translational velocity can be expressed by taking the time derivative of (\ref{tilde pd real}) as follows,
\begin{align}
\label{tilde dot pd real}
\dot{\tilde{p}}_d 
% & =\lambda_1 \begin{bmatrix} (\dot{\gamma} R  + \gamma  \dot{R} ) \cos(\delta)  - \gamma  R \omega_d \sin (\delta)  \\  (\dot{\gamma} R  + \gamma  \dot{R} )\sin(\delta) + \gamma  R \omega_d \cos (\delta) \\\lambda_1^{-1}\dot{\tilde{p}}_d^z   \end{bmatrix}\\
& ={\Bar{\lambda}_1}\begin{bmatrix}\dot \rho \cos (\delta)- \rho \dot \delta \sin (\delta) \\ \dot \rho \sin (\delta)+ \rho \dot \delta \cos (\delta)\\ \dot{\tilde{p}}_d^z  \end{bmatrix},
\end{align}
motivated by \cite{khoramshahi2018human}, $\dot{\rho}=k_\rho(\rho_d-\rho)$ represents the speed of convergence of $\rho$ to $\rho_d$,  with $k_\rho \in \mathbb R^+$ being a gain parameter. The desired angular velocity,  $\dot{\delta} \in \mathbb R^+$, can be selected by the user. 
% \begin{align}
%     R^2 = ({\Delta p^x})^2+({\Delta p^y})^2 \\
%     \dot{R} = \frac{1}{R} \left(\Delta p^x \frac{d}{dt} \Delta p^x + {\Delta p^y}\frac{d}{dt} {\Delta p^y} \right) \\
%     \dot{R} = \frac{1}{R}  \Delta p^T  \begin{bmatrix}
%         1 & 0 & 0\\
%         0 & 1 & 0\\
%         0 & 0 & 0
%     \end{bmatrix}\frac{d}{dt}\Delta p
% \end{align}
% As previously noted, the position along the $z$-axis is maintained constant at $\Bar{p}_d^z$. 
The desired velocity of the $z$--axis motion can consequently be generated using a  PD position controller expressed as, 
\begin{equation}
\label{dot tilde pdz}
\dot{\tilde{p}}_d^z = k_P ({\Bar{p}_d^z} - p^z) + k_D ({\dot{\Bar{p}}_d^z} - \dot{p}^z),
\end{equation}
where $k_P \in \mathbb R^+$ and $k_D \in \mathbb R^+$ represent the user-defined proportional and derivative gains, respectively. 

{The main task considered across three scenarios is tracing a circular path. The experiments include a non-interaction scenario (\hyperref[A]{Exp. A}), interaction with the EE  (\hyperref[B]{Exp. B}, \hyperref[D]{Exp. D}), and interaction with the robot body (\hyperref[C-1]{Exp. C} -- \hyperref[E]{Exp. E}). The non-interaction scenario serves as a baseline for evaluating tracking accuracy in 6D Cartesian space. In the EE interaction scenario, we examine the forces applied to the user while holding the EE, simulating a maintenance process. The robot body interaction scenario applies forces to the 4th joint along the $y$-axis relative to the base frame, mimicking external disturbances and environmental constraints. We experimentally compare our proposed method (\hyperref[C-2]{Exp. C--2}) with
the baseline (\hyperref[C-1]{Exp. C--1}), a classical approach (\hyperref[D]{Exp. D}), and two state-of-the-art interaction force estimation methods \cite{sadeghian2013task, lin2022unified} (\hyperref[E]{Exp. E}). Table \ref{tab2} summarizes the results.}
% These two evaluations investigate the impact of null space unknown interaction on Cartesian tracking accuracy, assessing the contribution of the proposed null space control method.

\subsection{Experiment A: Cartesian Path Tracking Without External Interaction}
\label{A}
In this experiment, we evaluate the Cartesian path tracking accuracy in a non-interaction scenario using the proposed method. Here, we set up a scenario where the Tool Center Point (TCP) cyclically traces a circular path with a 10 cm radius on the top surface of a cardboard box. Initially, the joint configuration is approximately at $[0, 2\pi/9, 0, -\pi/2, 0, 5\pi/18, 0]^T$ (rad) as shown in Fig. \ref{Experiment_Setup}, the EE is positioned at the circle center.  Additionally, the desired EE orientation $r_d$ is configured as $[\pi,0,\pi]^T$  (rad). This configuration ensures that the EE points vertically downward toward the surface. During motion, the yaw angle $\psi$ is constant relative to the base frame. Throughout the experiment, the robot remains free from external interaction forces, and the contact force along the $z$--axis of the EE is maintained close to zero, thus minimizing the effect of extraneous friction forces at the TCP. The control laws in  (\ref{dot pc})--(\ref{tauc}), (\ref{taun}), and (\ref{tilde dot pd real}), (\ref{dot tilde pdz}) are utilized in this experiment. The control parameters, detailed in Table \ref{tab1}, remain consistent across all experiments.
\begin{table}[H]
\centering
\caption{Control Parameters}
\renewcommand{\arraystretch}{1} % Adjust row height
\begin{tabular}{ | >{\centering\arraybackslash}m{1.5cm} | >{\centering\arraybackslash}m{2.5cm} | >{\centering\arraybackslash}m{2.5cm} | } 
  \hline
  Parameter & Value & Unit \\ 
  \hline
  $D$, $d_{n}$& $\operatorname{diag}(40, 40, 10)$, 5 & N$\cdot$ s$/$m \\ 
  \hline
  $K_{\omega}$ & $\operatorname{diag}(15, 15, 15)$ & N$\cdot$ m$/$rad \\ 
  \hline
  $\alpha_f$, $\alpha_d$ & 1 & -- \\ 
  \hline
  % $K_c$ & $\operatorname{diag}(30, 30, 10)$ & N$\cdot$ m$/$rad \\ 
  % \hline
  % $D_c$ & $\operatorname{diag}(20, 20, 10)$ & N$\cdot$ s$/$m \\ 
  % \hline
\end{tabular}
\label{tab1}
\end{table}

\begin{figure} [!ht]
   \begin{center}
\includegraphics[width=8.3cm,height=4.25cm]%width=7cm,height=3cm
{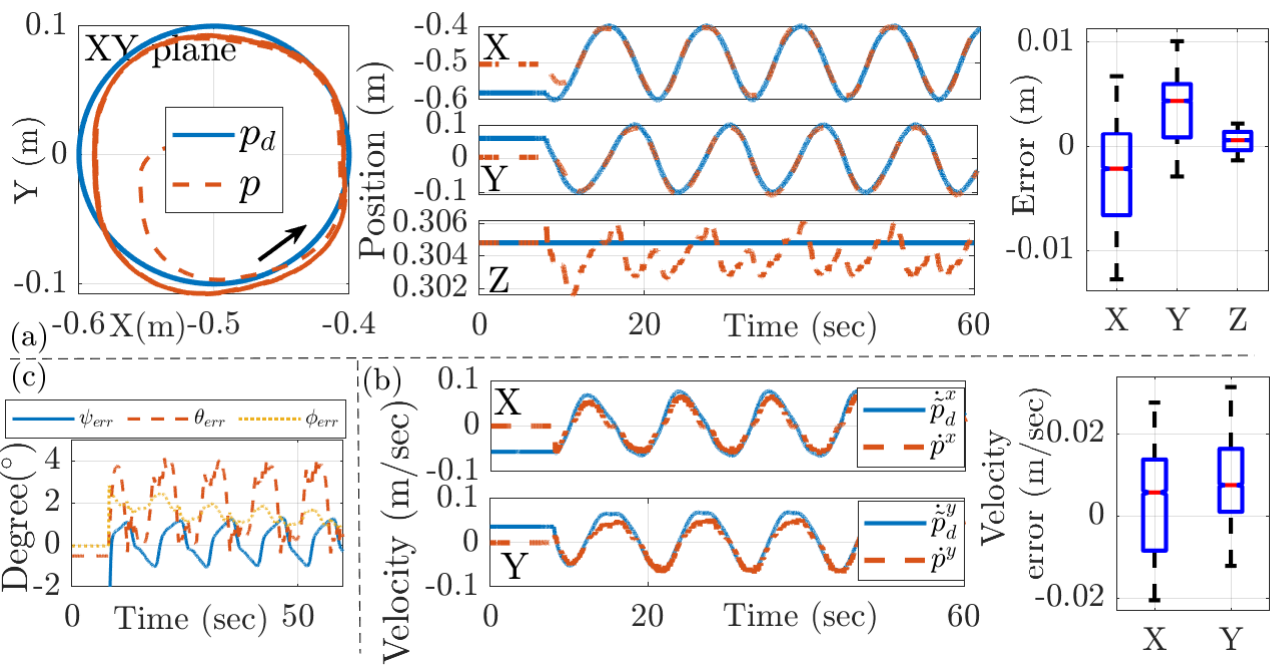}
   \end{center}
   \caption{Non-interactive (\hyperref[A]{Exp. A}): (a) Position, (b) velocity, and (c) orientation trackings for the proposed method.}
\label{no_contact_tracking+boxplot}
\end{figure}

Fig. \ref{no_contact_tracking+boxplot} (a) demonstrates tracking accuracy remains under 7 mm in the $XY$ plane, with height error along the $z$--axis remaining under 3 mm.  Minor fluctuations observed along the $z$--axis can be attributed to the uneven surface of the box.

% \begin{figure*} [t]
%    \begin{center}
% \includegraphics[width=0.95\textwidth,height=3cm]
% {Fig/no_contact_tracking+boxplot_dualcolumn_2.pdf}
%    \end{center}
%    \caption{Cartesian path tracking accuracy in a non-contact situation.}
% \label{no_contact_tracking+boxplot}
% \end{figure*}  
To ensure precise Cartesian path tracking, it is crucial to accurately track the generated desired velocity.
Therefore, it is essential to evaluate the accuracy of the Cartesian velocity tracking along both the 
$x$--axis and $y$--axis. As presented in Fig. \ref{no_contact_tracking+boxplot} (b), with a negligible error of approximately 1 cm$/$sec, the Cartesian velocity tracking controller demonstrates satisfactory performance, thereby contributing to the Cartesian position tracking accuracy.
% \begin{figure} [!ht]
%    \begin{center}
% \includegraphics[width=8.5cm,height=0.1cm]%width=7cm,height=3cm
% {Fig/no_contact_tracking_Vel_err_dualcolumn_2.eps}
%    \end{center}
%    \caption{Cartesian velocity tracking error.}
% \label{no_contact_tracking_Vel_err}
% \end{figure}  
To further evaluate the orientation stability of the EE during motion, we illustrate the orientation error in Fig. \ref{no_contact_tracking+boxplot} (c). 
The results demonstrate satisfactory performance, maintaining {orientation error within $ 4^{\circ}$. The cyclical pattern of the error aligns with the orientation control employing solely stiffness control, outlined in (\ref{Fcw}).  
% \begin{figure} [!ht]
% \begin{center}
% \includegraphics[width=8.5cm,height=0.1cm]{Fig/no_contact_rot_dualcolumn_2.eps}
% \end{center}
%    \caption{EE orientation error in non-contact situation.}
% \label{no_contact_rot}
% \end{figure}  

\subsection{Experiment B: Interaction on  End-Effector}
\label{B}
In this experiment, we investigate the interaction forces experienced by the user at the EE. The control laws applied are identical to those used in the preceding experiment.  Following the setup from the previous experiment, the scenario involves the TCP cyclically tracing a circular path with a radius of 10 cm. During the motion, the user holds the EE for about 10 seconds, repeating this action four times. The random interaction points are as indicated in Fig. \ref{EE_force_position}. 
Across four random interactions, the interaction forces range from 9 N to 18 N. The results demonstrate that the user can effortlessly execute the tool changeover procedure. In the last two interactions, the user demonstrates the ability to deviate the TCP from its desired path with slightly increased interaction forces.
\begin{figure} [t]
   \begin{center}
   \includegraphics[width=8cm,height=2.75cm]{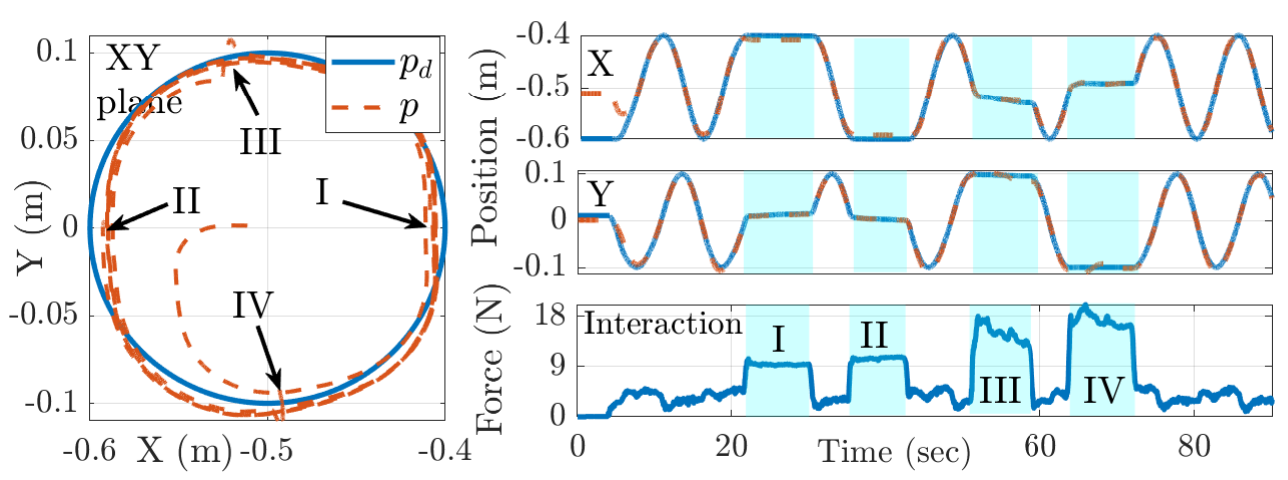}
   \end{center}
   \caption{{Interaction with the EE (\hyperref[B]{Exp. B}): position and measured interaction force using our proposed method. Shaded areas indicate periods of interaction.}}
\label{EE_force_position}
\end{figure}  

\subsection{{Comparative Experiment C--1: Interaction on Robot Body Without Proposed Null Space Control Method}}
\label{C-1}
In this series of comparative experiments, we investigate the efficacy of the null space impedance control scheme introduced in subsection \ref{Null Space Impedance Control}. We aim to demonstrate the proposed null space controller enables the redundant DOFs to respond to unknown external interactions compliantly, thereby reducing interference with Cartesian space motion.
% Thus, this experimental series contrasts the Cartesian path tracking accuracy under the effect of unknown external torques applied to the robot body in the case of \ref{C-1} and \ref{C-2}. 
\begin{figure} [t]
   \begin{center}
\includegraphics[width=8.5cm,height=2.1cm]{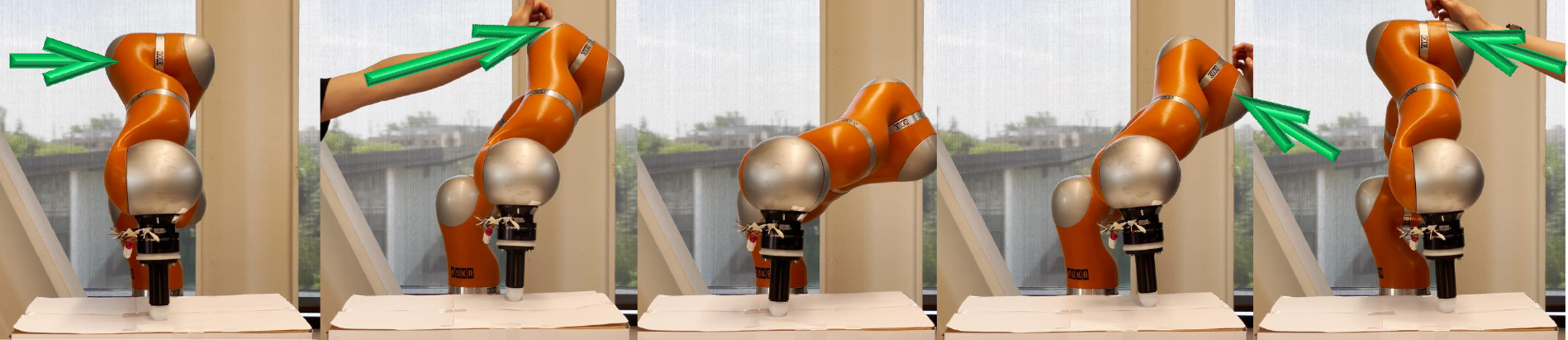}
   \end{center}
   \caption{Interaction directions (indicated by green arrows). }
\label{PUSH}
\end{figure} 

In the first experiment, the proposed null space control scheme is omitted. This allows the Cartesian motion to be controlled by the DS motion generator as described in \cite{khoramshahi2018human}, combined with the tracking controllers and Cartesian impedance controller used in the preceding experiments as a baseline approach. The initial joint configuration is the same as in \hyperref[A]{Exp. A}. During pre-interaction periods, the tracking accuracy deteriorates compared to Fig. \ref{no_contact_tracking+boxplot} (a).  At 32 and 61 seconds,  unknown interaction forces are applied on the 4$^{th}$ joint. The directions of the interactions are indicated in Fig. \ref{PUSH}. The duration of the interactions was approximately 2 and 4 seconds, respectively.
The impact of interactions on the EE path tracking is indicated in Fig. \ref{without_NS_Cartesian} (a) (labelled with I and II).  
As seen, these interactions led to noticeable deviations from the desired path. The error along the $z$--axis  ($\approx$ 2 cm) results in a loss of surface contact (Fig. \ref{without_NS_Cartesian} (b)). 
% This degradation is attributed to the absence of the proposed null space control scheme, highlighting the lack of null space impedance control and friction compensation for the robot joints.  Consequently, this results in increased deviations in Cartesian path tracking \cite{dietrich2015overview}.
% ficuciello2014cartesian

Furthermore, to demonstrate the behaviour of the robot body, we depict the variations in joint angles in Fig. \ref{without_NS_Cartesian} (c). 
The $q_1, q_2, q_3, q_5, q_6, q_7$ joints exhibit dynamic changes attributed to external interactions,  with joints $q_1, q_2, q_3, q_5, q_6$ being particularly affected. Note that the variation of $q_7$ is excluded as it only pertains to the change of orientation of the EE. The fluctuations in joints indirectly affected by external forces (such as $q_2$ and $q_6$) indicate that the joints directly affected by external forces are inefficient in deviating from their initial positions to dissipate external energy. This inefficiency causes more joints to be affected, ultimately impacting tracking accuracy.

\subsection{{Comparative Experiment C--2: Interaction on Robot Body With Proposed Null Space Control Method}}
\label{C-2}
This comparative experiment introduces the proposed null space control into the system. The control laws applied remain consistent with those in \hyperref[A]{Exp. A}. Pre-interaction periods exhibit enhanced tracking accuracy, closely matching the performance depicted in Fig. \ref{no_contact_tracking+boxplot} (a). We again apply unknown yet reasonably similar external forces ({justified in Table \ref{tab2})} to the 4$^{th}$ joint around the 33 and 62 seconds.  The interaction directions are labelled in Fig. \ref{PUSH}. The interaction duration aligns with \hyperref[C-1]{Exp. C--1}. The impact of interactions on the EE tracking is indicated in Fig. \ref{with_NS_Cartesian} (a) (labelled with I and II).
The Cartesian deviations induced by external interaction forces notably diminish compared to  Fig. \ref{without_NS_Cartesian} (a, b), which closely match the accuracy of \hyperref[A]{Exp. A}.
% These findings highlight the efficacy of the proposed null space control scheme in improving the Cartesian tracking accuracy.
\begin{figure} [t]
   \begin{center}
\includegraphics[width=8cm,height=3.5cm]{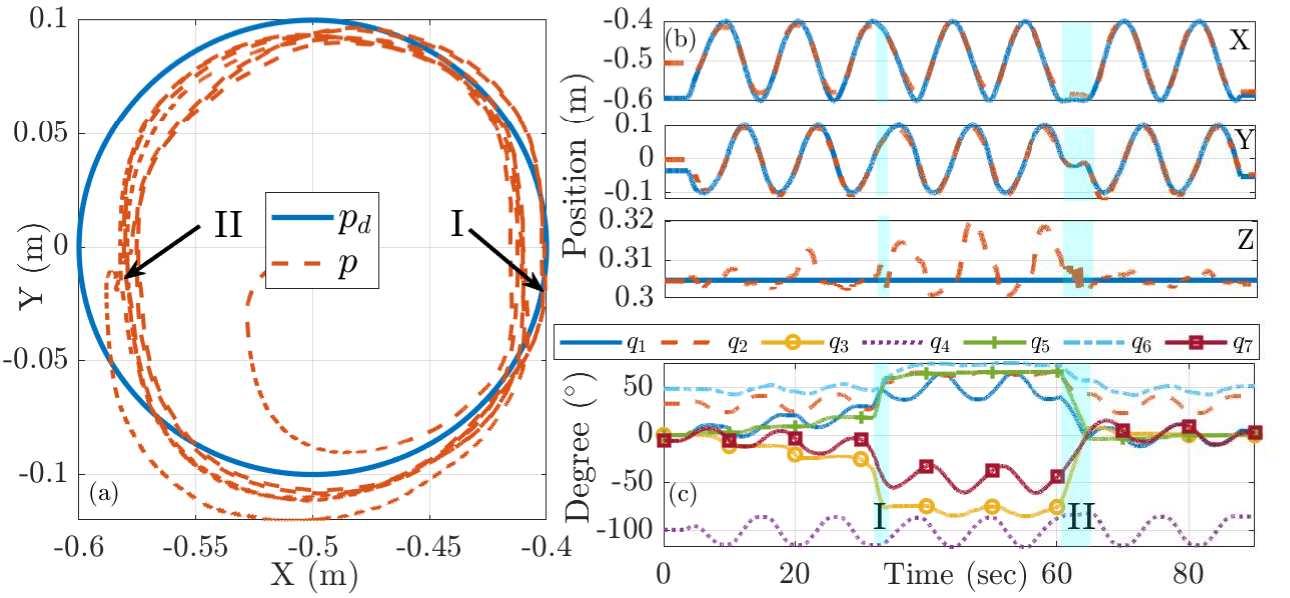}
   \end{center}
   \caption{Performance of the baseline method \cite{khoramshahi2018human} during robot body interaction (\hyperref[C-1]{Exp. C--1}): (a) Tracking performance, (b)  $x$--, $y$--, and $z$--axes variations, and (c) joint variations.}
\label{without_NS_Cartesian}
\end{figure}
\begin{figure} [t]
   \begin{center}
\includegraphics[width=8cm,height=3.5cm]{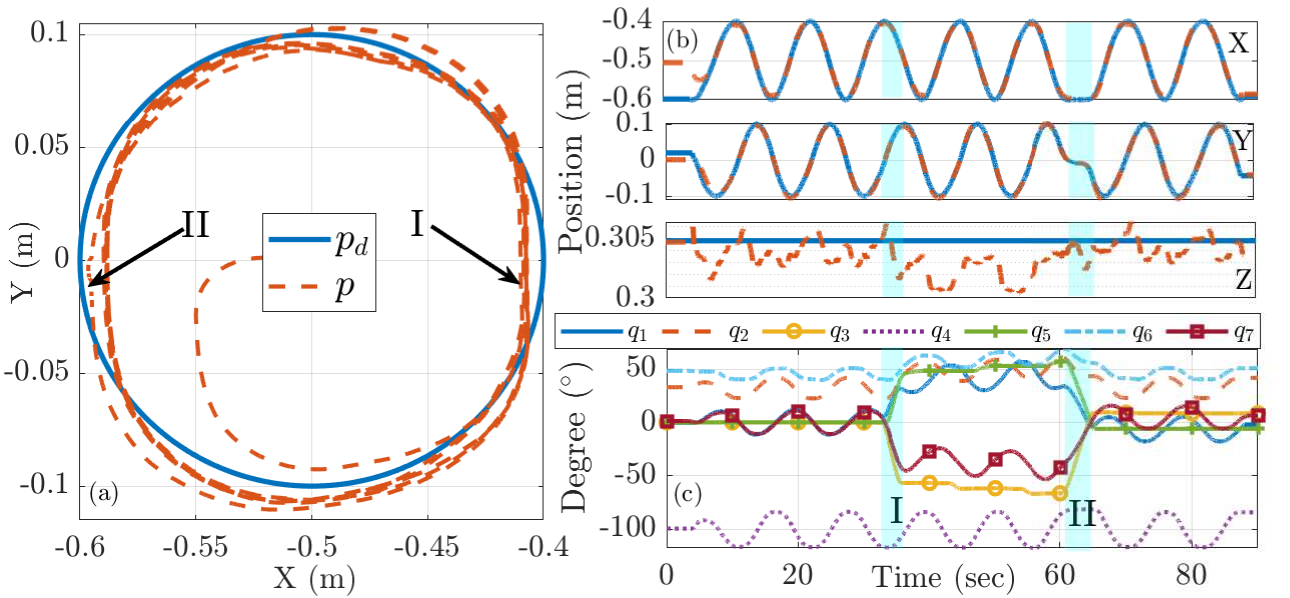}
   \end{center}
   \caption{Performance of the proposed method during robot body interaction (\hyperref[C-2]{Exp. C--2}): (a) Tracking performance, (b)  $x$--, $y$--, and $z$--axes variations, and (c) joint variations.}
\label{with_NS_Cartesian}
\end{figure}  

We further illustrate the variations in joint angles in Fig. \ref{with_NS_Cartesian} (c).
When the interaction force acts on the 4$^{th}$ joint,  only $q_1, q_3$, and $ q_5$ respond to the external forces, and again $q_7$ is excluded from the discussion.  This result demonstrates these joints compliantly {deviate ($\approx$50$^\circ$) from their initial positions, efficiently dissipating external energy within the null space. The robot's responsiveness is also enhanced by compensating for joint friction. Thus, fewer joints are affected, leading to the tracking accuracy remaining unaffected. 
% Note that the comparative experiments mentioned above only discussed changes in the $x$-axis and $y$-axis.  However, it is important to also account for changes in the $z$-axis, as they exhibit significant differences.
% Additionally, despite the initial joint configurations being the same in Figs. \ref{without_NS_Cartesian} and \ref{with_NS_Cartesian}, there are significant differences in joint angle variations during the initial non-contact periods.
% These additional findings are discussed in Section \ref{Discussion}.
\begin{figure} [!ht]
   \begin{center}
\includegraphics[width=8.5cm,height=6cm]{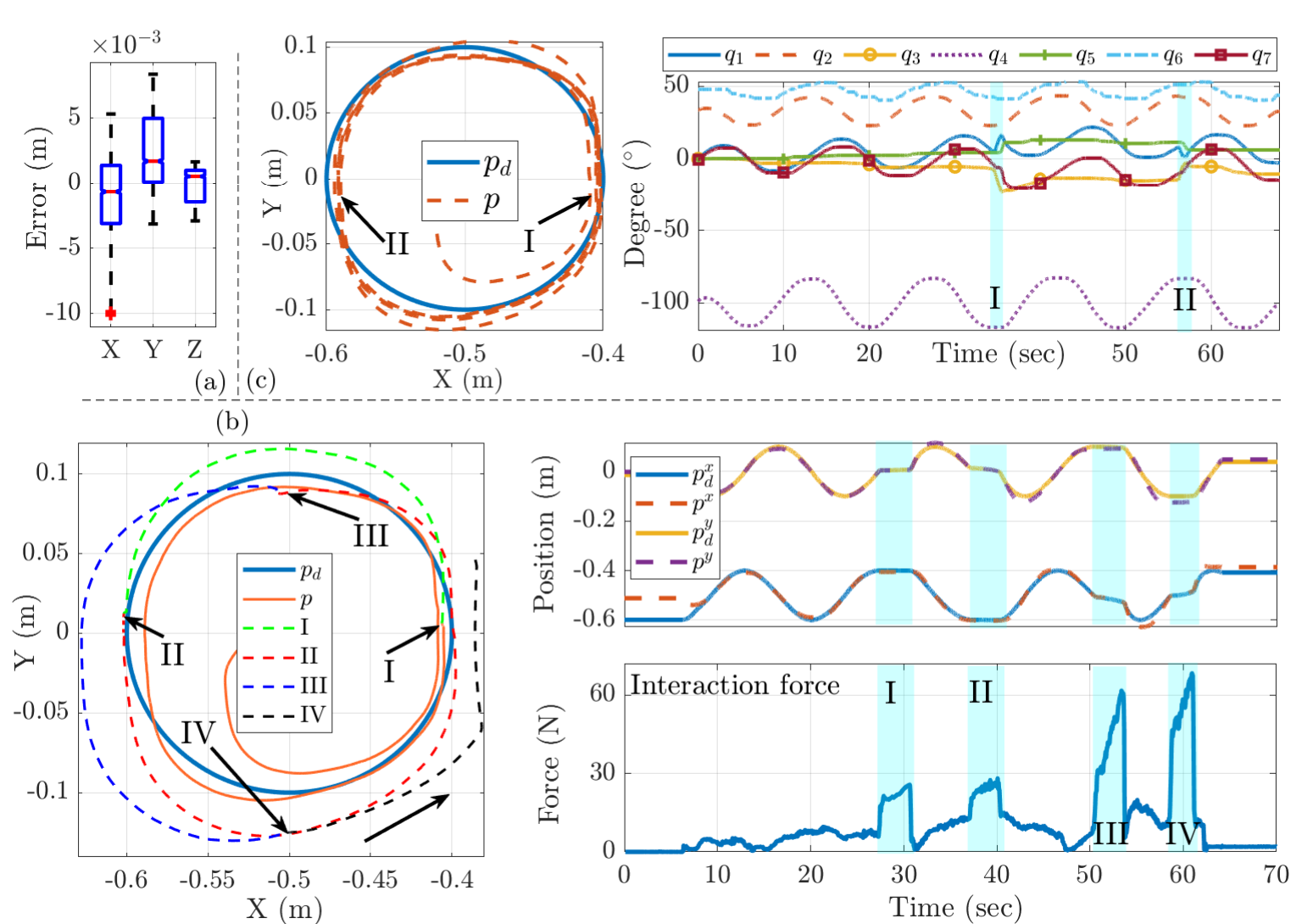}
   \end{center}
   \caption{{Performance of classical method \cite{hogan1985impedance} for comparison with (a) non-interaction (\hyperref[A]{Exp. A}), (b) EE interaction (\hyperref[B]{Exp. B}), and (c) robot body interaction (\hyperref[C-1]{Exp. C}).}}
\label{revise}
\end{figure}

% \begin{figure} [!ht]
%    \begin{center}
% \includegraphics[width=8.5cm,height=6cm]{Fig/Stateofart_updateRLSE.eps}
%    \end{center}
%    \caption{\textcolor{red}{Performance evaluation of (a) Observer method \cite{sadeghian2013task}, (b) RLSE method \cite{lin2022unified} during interaction with the robot body. (1) Position tracking, (2) joint variations, (3) pre-interaction tracking errors, and (4) z-axis variations.}}
% \label{stateofart}
% \end{figure}
\begin{figure} [!ht]
   \begin{center}
\includegraphics[width=8.7cm,height=2.9cm]{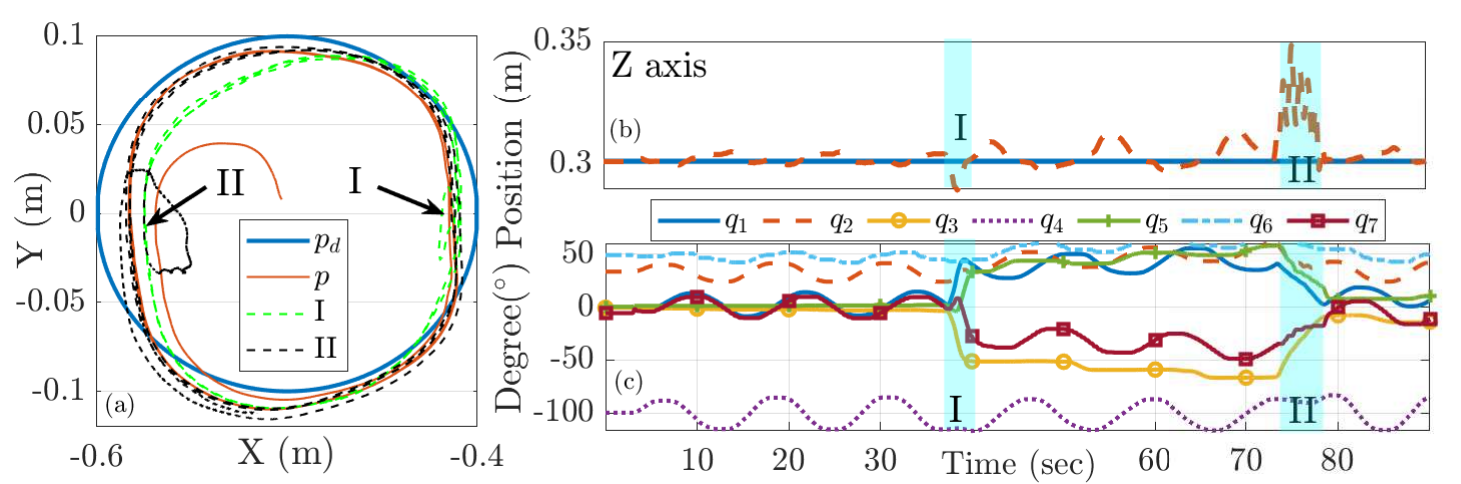}
   \end{center}
   \caption{{Performance of the Observer method \cite{sadeghian2013task} during robot body interaction (\hyperref[E]{Exp. E}): (a) Tracking performance, (b) $z$--axis variations, and (c) joint variations.}}
\label{stateofart_Observer}
\end{figure}
\begin{figure} [!ht]
   \begin{center}
\includegraphics[width=8.5cm,height=2.8cm]{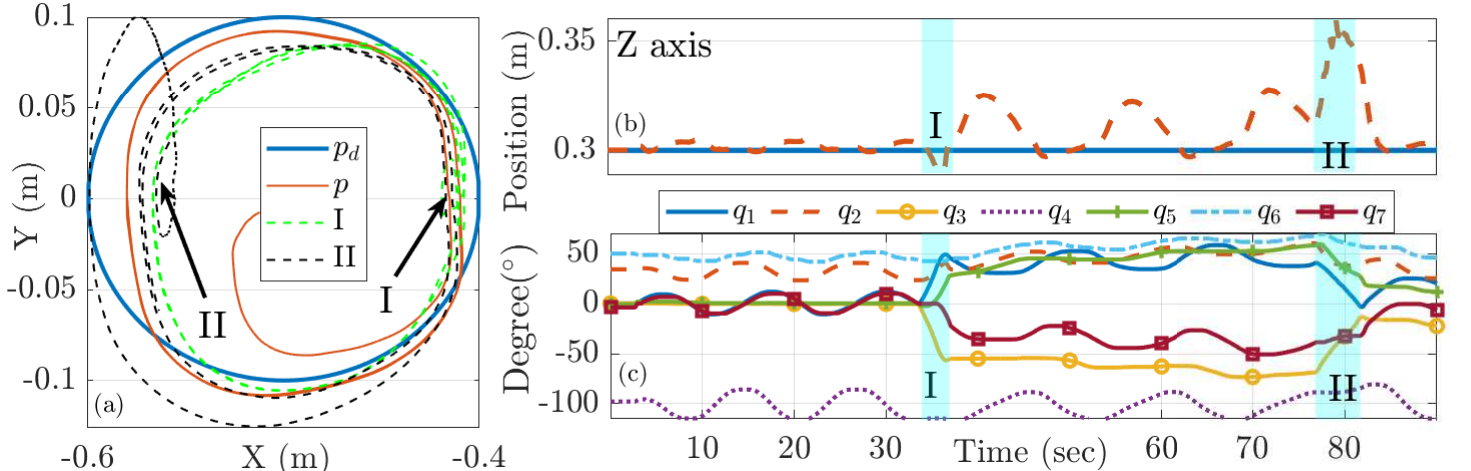}
   \end{center}
   \caption{{Performance of the RLSE method \cite{lin2022unified} during robot body interaction (\hyperref[E]{Exp. E}): (a) Tracking performance, (b) $z$--axis variations, and (c) joint variations.}}
\label{stateofart_RLSE}
\end{figure}
% \begin{table*}[t]
% \centering
% \caption{\textcolor{red}{Comparison of Null Space Interaction Forces and Tracking Accuracy Across $x$-, $y$-, and $z$-axes Directions}}
% \renewcommand{\arraystretch}{1.0} % Reduce row height
% \setlength{\tabcolsep}{12pt} % Adjust column spacing for better fit
% \begin{tabular}
% {|c|c|c|c|c|c|}
%   \hline
%    & \multicolumn{5}{c|}{Methods} \\ 
%    \cline{2-6}
%    & Classical\cite{hogan1985impedance} & Baseline \cite{khoramshahi2018human} & \textcolor{red}{Proposed} & Observer \cite{sadeghian2013task} & RLSE \cite{lin2022unified} \\ 
%   \hline
%   Max Force, I / II (N) & 47/47 & 35/30 & 29/26 & 36/30 & 40/41 \\ 
%   \hline
%   Pre-interaction NMAE, X/Y/Z (\%) & 5.0/4.2/1.5 & 10.3/6.0/2.2 & 6.4/4.5/1.0 & 8.8/5.4/1.3 & 10.1/4.7/1.4 \\ 
%   \hline
%   Post-interaction NMAE, X/Y/Z (\%) & 5.1/8.7/1.5 & 12.3/11.0/5.7 & 5.9/6.2/1.1 & 23.0/11.0/15.8 & 22.0/14.4/19.7 \\ 
%   \hline
% \end{tabular}
% \label{tab2}
% \end{table*}
\begin{table*}[t]
\centering
\caption{{Comparison of Interaction Forces on the Robot Body, Joint Deviations, and Tracking Accuracy.}}
\renewcommand{\arraystretch}{1.0} % Reduce row height
\setlength{\tabcolsep}{12pt} % Adjust column spacing for better fit
\begin{tabular}
{|c|c|c|c|c|c|}
  \hline
   & Classical\cite{hogan1985impedance} & Baseline \cite{khoramshahi2018human} & \textbf{Proposed} & Observer \cite{sadeghian2013task} & RLSE \cite{lin2022unified} \\ 
  \hline
  Max force, I / II (N) & 47/47 & 35/30 & \textbf{29/26} & 36/30 & 40/41 \\ 
  \hline
  Joint deviation, I $\&$ II ($^\circ$) & $\approx$10 & $\approx$45 & $\approx$\textbf{50} & $\approx$45 & $\approx$41 \\ % Replace "XX" with actual values
  \hline
  Pre-interaction NMAE, X/Y/Z (\%) & \textbf{5.0}/\textbf{4.2}/1.5 & 10.3/6.0/2.2 & 6.4/4.5/\textbf{1.0} & 8.8/5.4/1.3 & 10.1/4.7/1.4 \\ 
  \hline
  Post-interaction NMAE, X/Y/Z (\%) & \textbf{5.1}/8.7/1.5 & 12.3/11.0/5.7 & 5.9/\textbf{6.2}/\textbf{1.1} & 23.0/11.0/15.8 & 22.0/14.4/19.7 \\ 
  \hline
\end{tabular}
\label{tab2}
\end{table*}

\subsection{Comparative experiment D: Classical Impedance Control}
\label{D}
For a thorough evaluation, we compare a classical impedance control method presented in \cite{hogan1985impedance} with all prior experiments.
% , which has since undergone various improvements \cite{ haddadin2024unified,dimeas2016manipulator}. 
The control law is given by $K_c(p_d - p)+D_c(\dot{p}_d-\dot{p})$, where $K_c = \operatorname{diag} (30,30,30)$ and $D_c = D$ represent the stiffness and damping matrices, respectively, selected to be consistent with the range used in previous experiments.  EE orientation is controlled by (\ref{Fcw}).  
Fig. \ref{revise} (a) demonstrates Cartesian path tracking accuracy in the non-contact scenario as in \hyperref[A]{Exp. A}. The results indicate a similar level of tracking accuracy as in Fig. \ref{no_contact_tracking+boxplot} (a). To maintain brevity, only the error signal boxplot is presented. However, when physical interactions are introduced, as in \hyperref[B]{Exp. B} and \hyperref[C-1]{Exp. C} the robot's performance significantly degrades, as shown in Fig. \ref{revise} (b) and (c), respectively. In Fig. \ref{revise} (b), \hyperref[B]{Exp. B} is repeated, but the forces required to hold the tool in the current position increase sharply, by approximately three times, reaching a maximum of 60 N, which results in less interaction time ($\approx$ 3 sec). The interaction forces can continue to rise if the user does not release the tool. 
% The post-interaction motion \textcolor{blue}{deviates substantially from the desired path by approximately 25$\%$ }compared to Fig. \ref{EE_force_position}.
In Fig. \ref{revise} (c), \hyperref[C-1]{Exp. C} is repeated. 
% Although the applied forces remain consistent with those in Experiment C and 
% Due to the stiffness term, Cartesian deviations are ignorable, however, the interaction forces applied by the user are higher, and \textcolor{blue}{joints fail to deviate (approximately only 10$^\circ$)} in response to external forces. This restricts the ability to dissipate the external energy applied to the robot body. As a result, this leads to less compliance and more rigid behaviour during the interactions.
The stiffness term reduces Cartesian deviations, but increases user-applied interaction forces {(as justified in Table \ref{tab2})}. Joint deviations are limited ($\approx$$10^\circ$), the dissipation of external energy is therefore inefficient and results in less compliance and a more rigid behaviour.

\subsection{Comparative Experiment E: Interaction on Robot Body with Null Space Interaction Estimation Methods}
\label{E}
{This section compares null space interaction handling performance using two interaction force estimation methods: a momentum-based observer (Fig. \ref{stateofart_Observer}) from \cite{sadeghian2013task} and recursive least square estimation (RLSE) (Fig. \ref{stateofart_RLSE}) from \cite{lin2022unified}. The Cartesian control laws follow those in \hyperref[A]{Exp. A}, but the estimated null space interaction forces replace the proposed $\tau^n$ in \eqref{taun}.
While the interaction directions match those in \hyperref[C-1]{Exp. C}, a slightly longer duration is needed to achieve comparable joint deviation due to increased resistance.  Both methods exhibit similar tracking accuracy pre-interaction but fail to maintain main task tracking during and after interactions, particularly with significant $z$-axis errors ($\approx$5 cm), resulting in a loss of surface contact.
}

{Table \ref{tab2} provides a numerical comparison of the five methods, evaluating maximum interaction forces during two interactions, interaction-induced joint deviations, and the normalized maximum absolute error (NMAE), given by \eqref{NMAE} along the $x$--, $y$--, and $z$--axes. Maximum interaction forces vary between 26 N and 47 N due to differences in null space compliance, with nearly identical forces observed across the two interactions for each method.
Joint deviations, averaged for $q_3$ and $q_5$ across interactions, show similar values across methods (except for the classical method), indicating consistent interaction scenarios despite variations in resistance. The proposed method effectively dissipates external energy through redundant DOFs, reducing resistance for the user. In contrast, the classical method, being relatively rigid, offers accurate tracking but limited deviation.
Pre-interaction tracking exhibits a mere 5$\%$ difference among methods, with the classical and proposed methods performing best. However, during and after interactions, the tracking accuracy of other methods degrades, with errors increasing by up to 18$\%$ along the $x$-- and $z$--axes, highlighting the effectiveness and superiority of the proposed method.
}

\section{{Discussion and Conclusion}}
\label{Discussion}
{The comparative results are satisfactory.  The results in Table \ref{tab2} represent the averages from multiple trials.}
% yet it is crucial to maintain the joint velocities induced by external forces close to the robot's upper hardware limits to ensure the widest possible range of forces can be safely covered.
% maintain the interaction torques within an upper boundary. This precaution is critical to avoid triggering software or hardware joint limits/interruptions.  
The efficacy of the null space control law relies on the degrees of redundancy. Due to the constraints imposed by the main task,  forces along the $y$--axis of the base frame and at or around the 4$^{th}$ joint of the robot could be dissipated in the null space. Slight variations in maximum interaction forces reflect human's variability. {To avoid joint limit violations and singularities from aggressive interaction forces, the joint angles were kept within an optimal range. } Future work will extend the permissible EE and robot body interaction forces while avoiding singularities and joint limits \cite{haddadin2024unified},\cite{dimeas2016manipulator}. 
% One potential solution involves improving the performance of the velocity tracking controller. 
% Moreover, the control parameters investigated in this study are set as fixed values based on empirical trials to align with the robot's observed performance. Nevertheless, there is potential to adjust these parameters using adaptive control to accommodate various interaction scenarios.

%\section{Conclusion}
%\label{Conclusion}
This study introduced a control approach that enabled compliant interaction at both the EE and the robot body while maintaining main task tracking accuracy. The proposed method did not rely on measuring or estimating external interaction forces, nor on robot dynamics information. Its passivity was proven even in the presence of unknown interaction forces and robot dynamics. Comprehensive experimental validation on a KUKA LWR IV+ robot confirmed its efficacy and practical feasibility.

\bibliographystyle{ieeetr}
\bibliography{bib}
\end{document}